\documentclass[preprint,12pt]{elsarticle}

\usepackage{amssymb}
\usepackage{amsmath}
\usepackage{bm}
\usepackage{algorithmic}
\usepackage{algorithm}
\usepackage{subfig}
\usepackage{booktabs}
\usepackage{multirow}
\usepackage{newtxtext}
\usepackage{makecell}
\usepackage{url}
\usepackage[colorlinks=false]{hyperref}
\usepackage{geometry}
\usepackage{color}			
\journal{}

\begin{document}

\begin{frontmatter}



\title{An Adaptive Dropout Approach for \\ High-Dimensional Bayesian Optimization}


\author{Jundi~Huang}

\author{Dawei~Zhan\corref{cor}}
\ead{zhandawei@swjtu.edu.cn}
\cortext[cor]{Corresponding author}

\affiliation{organization={School of Computing and Artificial Intelligence, Southwest Jiaotong University},
            city={Chengdu},
            country={China}}

\begin{abstract}
Bayesian optimization (BO) is a widely used algorithm for solving expensive black-box optimization problems. However, its performance decreases significantly on high-dimensional problems due to the inherent high-dimensionality of the acquisition function.
In the proposed algorithm, we adaptively dropout the variables of the acquisition function along the iterations. By gradually reducing the dimension of the acquisition function, the proposed approach has less and less difficulty to optimize the  acquisition function.
Numerical experiments demonstrate that AdaDropout effectively tackle high-dimensional challenges and improve solution quality where standard Bayesian optimization methods often struggle. Moreover, it achieves superior results when compared with state-of-the-art high-dimensional Bayesian optimization approaches. This work provides a simple yet efficient solution for high-dimensional expensive optimization.

\end{abstract}



\begin{keyword}
Bayesian optimization \sep Expensive optimization \sep  High-dimensional optimization \sep Adaptive Dropout.

\end{keyword}

\end{frontmatter}


\section{Introduction}
\label{sec_introduction}
Bayesian optimization (BO) \cite{Mockus_1975, Mockus_1994}, also known as efficient global optimization (EGO) \cite{Jones_1998}, is an effective optimization technique commonly used to solve expensive black-box problems. By employing a Gaussian Process model \cite{Rasmussen_2006} to approximate the objective function, Bayesian optimization can predict the behavior of the function at untested points, thus reducing the number of expensive evaluations. The selection of new samples is guided by an acquisition function that balances exploration and exploitation. Common acquisition functions include expected improvement, lower confidence bound, and probability of improvement \cite{Jones_2001}. The use of Gaussian Process models and carefully designed acquisition functions makes Bayesian optimization an extremely sample-efficient method \cite{Snoek_2012, Shahriari_2016, Guo_2021}. It has been successfully applied to a wide range of optimization problems, including multi-objective optimization \cite{Zhang_2010, Yang_2019, Han_2022} and parallel optimization \cite{Briffoteaux_2020, Chen_2023, Zhan_2023, WangZ_2023}.

However, when addressing high-dimensional problems, Bayesian optimization faces challenges due to the increased complexity of the search space \cite{Shahriari_2016}, which requires more samples to maintain the accuracy of the surrogate model. In high-dimensional cases, the acquisition function often exhibits highly nonlinear and multimodal behavior, with many local optima, which might trap the optimization algorithm. This results in a significant decline in the sample efficiency of Bayesian optimization, as the algorithm may waste evaluations in suboptimal regions of the search space. Over the past two decades, various strategies, such as variable selection, low-dimensional embeddings, additive models and local Gaussian process models have been proposed to address these challenges \cite{Binois_2022}.

The first type of approaches reduces the complexity of the high-dimensional problems through variable selection. 
The Dropout approach \cite{Li_2017} optimizes only a randomly chosen subset of dimensions at each iteration while infilling the remaining variables using three strategies: random sampling, copying values from the best historical solution, or a hybrid approach.
The Monte Carlo Tree Search (MCTS) method \cite{Song_2022} dynamically partitions variables into important and unimportant subsets, iteratively refining the search space based on historical evaluations, thereby improving optimization efficiency. 
Other effective variable selection techniques include posterior-based filtering \cite{Linkletter_2006}, hierarchical diagonal sampling combined with likelihood ratio testing \cite{Chen_2012}, and indicator-based methods \cite{Zhang_2023}, which further refine selection criteria. Additionally, dimensionality scheduling algorithms \cite{Ulmasov_2016} reduce computational overhead by optimizing only a subset of dimensions at each iteration, facilitating faster convergence.

Another approach to improving High-Dimensional Bayesian Optimization (HDBO) is through low-dimensional subspace embeddings. 
The Random Embedding Bayesian Optimization (REMBO) algorithm \cite{Wang_2016} projects high-dimensional problems into randomly selected low-dimensional subspaces using a Gaussian random matrix. It then performs Bayesian optimization in the embedded space and maps the solutions back to the original space, which  effectively reduces computational complexity.
The Bayesian Optimization with Adaptively Expanding Subspaces (BAxUS) approach \cite{papenmeier_2022} starts with a small subspace and gradually increasing its dimensionality over iterations, improving the probability of containing the global optimum. Other methods, such as improved kernel functions \cite{Binois_2015}, sequential random embedding \cite{Qian_2016}, partial least squares with Kriging models \cite{Bouhlel_2016}, sliced inverse regression \cite{Zhang_2019}, and hashing-enhanced subspace optimization \cite{Nayebi_2019}, further refine subspace selection and computational efficiency. However, these embedding-based methods assume a low-dimensional structure, which may not always exist. They also struggle with boundary distortions. These limitations reduce their effectiveness in complex real-world scenarios.

Furthermore, various methods have been proposed to solve HDBO problems using additive models. The Additive Gaussian Process Upper Confidence Bound (Add-GP-UCB) algorithm \cite{Kandasamy_2015} reduces the complexity of the high-dimensional problem by assuming that the objective function can be decomposed into a sum of lower-dimensional sub-functions, each depending only on a small subset of variables. This structure enables independent Gaussian process models for each sub-function, allowing optimization to be performed in smaller, computationally manageable subspaces.  Extensions of this idea include projection-based additive models \cite{Li_2016}, optimization with overlapping additive groups \cite{Rolland_2018} and structured kernel learning with batch evaluation \cite{Wang_2017}. Despite their advantages, additive model-based approaches rely on the assumption that the objective function can be decomposed into meaningful low-dimensional components. If this assumption does not hold, the performances of these algorithms might degrade.

In addition, local Gaussian Process (GP) models provide an alternative strategy by focusing on trust regions and adaptive search mechanisms to balance exploration and exploitation efficiently. The Trust Region Bayesian Optimization (TuRBO) algorithm \cite{Eriksson_2019} addresses the limitations of global surrogate modeling by maintaining multiple trust regions, each with an independent local GP model. It employs a multi-armed bandit strategy to allocate samples across these trust regions dynamically, ensuring efficient exploration while refining promising areas. The Taking-Another-Step Bayesian Optimization (TAS-BO) \cite{GUI_2024} further enhances local search by first identifying a candidate solution using a global GP and then training a local GP around it to refine the selection, improving search precision in high-dimensional spaces. 

Finally, beyond these mainstream approaches, several other methods have been developed. The Coordinate Line Bayesian Optimization (CoordinateLineBO) \cite{Kirschner_2019} improves the scalability of Bayesian optimization by decomposing the acquisition function into a sequence of one-dimensional subproblems. Unlike traditional high-dimensional BO methods, CoordinateLineBO restricts the search to carefully selected one-dimensional subspaces containing the best-known point, allowing for efficient exploration without solving high-dimensional acquisition function optimization. The Bayesian optimization with cylindrical kernels method \cite{Oh_2018} transforms the search space using cylindrical kernels. This improves kernel behavior and addresses fixed-point dependency issues. 

Many existing approaches in Bayesian optimization make structural assumptions about the objective function being optimized. These assumptions can significantly limit the performance of the algorithms when the underlying function does not align with the assumed model.

In this work, we propose the Adaptive Dropout approach (AdaDropout) for high-dimensional Bayesian optimization, which does not make structural assumptions about the problem to be optimized. The key contributions of our work are as follows:
\begin{enumerate}
	\item The proposed approach dynamically drops out variables of the acquisition function during the optimization process. By dynamically adjusting the number of optimizing variables, the proposed approach is able to achieve a better balance between exploration and exploitation.
	\item We conduct numerical experiments to evaluate the performance of AdaDropout and compare it with the standard BO and six state-of-the-art high-dimensional Bayesian optimization algorithms. The experimental results demonstrate the superiority of our approach in solving high-dimensional expensive optimization problems, in both computational cost and optimization efficiency.
\end{enumerate}

The rest of this paper is organized as follows. Section~\ref{section_background} introduces the backgrounds about the Gaussian process model and the Bayesian optimization algorithm.  Section~\ref{section_coordinate_descent} describes the adaptive dropout based Bayesian optimization algorithm. Section~\ref{section_experiment} presents the corresponding numerical experiments. Conclusions about this paper are given in Section~\ref{section_conclusion}.

\section{Backgrounds}
\label{section_background}

This work considers a single-objective black-box optimization problems, where the objective function is expensive to evaluate. Formally, consider a noise-free objective function \(f: \mathcal{X} \subseteq \mathbb{R}^D \to \mathbb{R}\), where \(\mathcal{X}\) is the input space, and we aim to find the global minimum
\begin{equation}
\bm{x}^* = \underset{\bm{x} \in \mathcal{X}}{\mathrm{argmin}} \, f(\bm{x})
\end{equation} 
where \(\bm{x}^*\) represents the optimum. Bayesian optimization is particularly valuable for problems where evaluations of \(f(\bm{x})\) are costly. The framework seeks to minimize the number of function evaluations required by efficiently guiding the search for the optimum.

The Bayesian optimization process comprises two core components: Gaussian Process (GP) model serving as a probabilistic surrogate model for the objective function and acquisition function leveraging the GP model to balance exploration and exploitation.

\subsection{Gaussian Process Model}
The Gaussian Process (GP) model is widely used in Bayesian optimization, providing a flexible and probabilistic framework for surrogate modeling. A GP defines a distribution over functions, where any finite set of points follows a multivariate Gaussian distribution. The predictive mean and variance of the GP provide not only function value estimates but also uncertainty quantification, enabling the efficient selection of evaluation points.

In a Gaussian Process (GP) model, the correlation function defines the dependence structure between function values at different input points. Commonly used correlation functions include the Radial Basis Function (RBF) correlation, the Periodic correlation, and the Linear correlation. In this work, we use the RBF correlation function, which is defined as \cite{Rasmussen_2006}

\begin{equation}
	r(\bm{x_i}, \bm{x_j}) = \exp\left[-\frac{1}{2l^2} (\bm{x_i} - \bm{x_j})^T(\bm{x_i} - \bm{x_j})\right]
\end{equation}

where \(l\) is the length-scale hyperparameter that controls the smoothness of the function. This hyperparameter can be tuned by maximizing the likelihood of the observed data.

For a dataset of \(n\) observed points \(\{\bm{x}^{(i)}, f^{(i)}\}_{i=1}^n\), the predictive mean and variance at a new input point \(\bm{x}\) are given by \cite{Jones_1998}

\begin{equation}
\hat{y}(\bm{x}) = \hat{\mu} + \bm{r}^T \bm{R}^{-1} (\bm{f} - \bm{1} \hat{\mu})
\label{GP_mean}
\end{equation}

\begin{equation}
\hat{s}^2(\bm{x}) = \hat{\sigma}^2 \left[ 1 - \bm{r}^T \bm{R}^{-1}\bm{r} + \frac{(1 - \bm{1}^T \bm{R}^{-1} \bm{r})^2}{\bm{1}^T \bm{R}^{-1} \bm{1}} \right]
\label{GP_variance}
\end{equation}
with
\begin{equation}
\hat{\mu} = \frac{\bm{1}^T \bm{R}^{-1} \bm{f}}{\bm{1}^T \bm{R}^{-1} \bm{1}}
\end{equation}

\begin{equation}
\hat{\sigma}^2 = \frac{(\bm{f} - \bm{1} \hat{\mu})^T \bm{R}^{-1} (\bm{f} - \bm{1} \hat{\mu})}{n}
\end{equation}
where \(\hat{\mu}\) and \(\hat{\sigma}^2\) are the mean and the variance of the GP, \(\bm{r} = [r(\bm{x}, \bm{x}^{(1)}), \cdots, r(\bm{x}, \bm{x}^{(n)})]^T\) is the vector of correlation coefficients between \(\bm{x}\) and the observed points, $\bm{R}$ is the \(n \times n\) correlation matrix of the observed points, with entries \(R_{ij} = r(\bm{x}^{(i)}, \bm{x}^{(j)})\), \(\bm{f} = [f^{(1)}, \cdots, f^{(n)}]^T\) is the vector of observed function values and $\bm{1}$ is a vector of ones.

The predictive mean \(\hat{y}(\bm{x})\) reflects the GP's best estimate of the function value at \(\bm{x}\), while the predictive variance \(\hat{s}^2(\bm{x})\) quantifies the uncertainty in that prediction. This combination of mean and variance makes the GP an effective surrogate model for guiding exploration and exploitation in Bayesian optimization.

\begin{figure}
	\centering
	\includegraphics[width=0.6\linewidth]{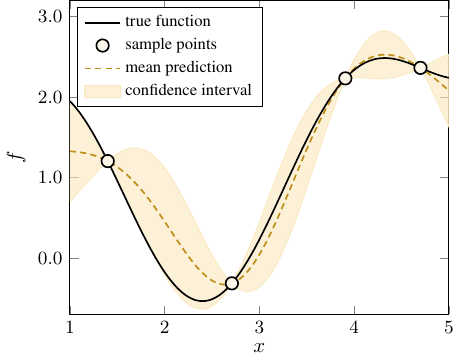}
	\caption{GP approximation of the $f = \cos(x)+\sin(2x)+0.5x$ function }
	\label{fig_GP}
\end{figure}

The GP model applied to a one-dimensional function $f = \cos(x)+\sin(2x)+0.5x$ is shown in Fig.~\ref{fig_GP}. In this figure, the solid black line represents the true function, the circles denote the observed sample points used to fit the GP model, the dashed brown line corresponds to the GP model's predictive mean and the shaded area around the dashed line represents the confidence interval. Those behaviors on the figure highlight a key feature of GPs, their ability to quantify uncertainty in predictions. This is critical in guiding the exploration-exploitation trade-off in Bayesian optimization.

\subsection{Acquisition Function}
Popular acquisition functions in Bayesian optimization include the Expected Improvement (EI) ~\cite{Mockus_1975,Mockus_1994}, probability of improvement ~\cite{Jones_2001} and lower confidence bound ~\cite{Jones_2001}. Among these, EI is arguably the most widely used due to its performance remarkably well in striking a balance between exploration and exploitation.

\begin{figure}
	\centering
	\includegraphics[width=0.6\linewidth]{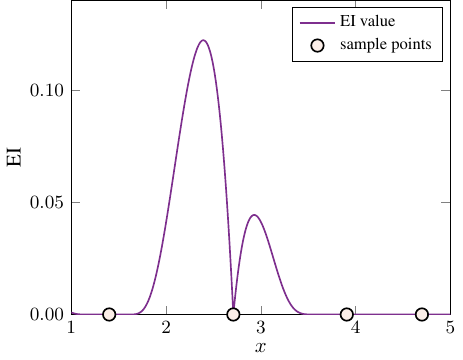}
	\caption{The EI function on the $f = \cos(x)+\sin(2x)+0.5x$ problem}
	\label{fig_EI}
\end{figure}

For a candidate point \(\bm{x}\), the corresponding objective value is modeled as a random variable following a normal distribution \(\mathcal{N}(\hat{y}(\bm{x}), \hat{s}^2(\bm{x}))\), where \(\hat{y}(\bm{x})\) and \(\hat{s}^2(\bm{x})\) are the predictive mean and variance of the Gaussian Process (GP), as defined in Equations~(\ref{GP_mean}) and (\ref{GP_variance}), respectively.

The Expected Improvement is defined as the expected value of the improvement \(I(\bm{x})\), where \cite{Jones_1998} 
\begin{equation}
I(\bm{x}) = \max(f_{\text{min}} - Y, 0)
\end{equation} 
with \(f_{\text{min}}\) representing the best observed function value so far and \(Y\) being a realization of the random variable at \(\bm{x}\). The EI acquisition function can then be expressed analytically as \cite{Jones_1998}
\begin{align}
	\text{EI}(\bm{x}) &= \mathbb{E} \left[ \max \left( f_{\min} - Y, 0 \right) \right] \notag \\
	&= \left( f_{\text{min}} - \hat{y}(\bm{x}) \right) \Phi \left( \frac{f_{\text{min}} - \hat{y}(\bm{x})}{\hat{s}(\bm{x})} \right)
	+ \hat{s}(\bm{x}) \phi \left( \frac{f_{\text{min}} - \hat{y}(\bm{x})}{\hat{s}(\bm{x})} \right)
\end{align}

where \(\Phi(\cdot)\) is the cumulative distribution function of the standard normal distribution, and \(\phi(\cdot)\) is its probability density function.  

Intuitively, EI balances the trade-off between exploration and exploitation. The first term captures the improvement weighted by its probability, while the second term represents the contribution of uncertainty in unexplored regions. 
The EI function derived from the GP model for a one-dimensional objective function is illustrated in Fig.~\ref{fig_EI}. As shown in the figure, the EI function exhibits a multi-modal structure. Its value is exactly zero at the sample points. Between these sample points, the EI function rises, indicating regions where potential improvements are expected.

\subsection{Bayesian Optimization}

Bayesian optimization (BO) is an efficient optimization framework for expensive black-box functions. The process consists of two main components: building a probabilistic surrogate model, commonly a Gaussian Process (GP), to approximate the objective function, and optimizing an acquisition function to locate a new point, such as Expected Improvement (EI), to decide the next sampling point.

\begin{algorithm}
	\caption{Computational Framework of BO}
	\label{algorithm_BO}
	\begin{algorithmic}[1]	
		\REQUIRE $N_{\text{init}}$: number of initial samples; 
		$N_{\text{max}}$: maximum allowed function evaluations.
		\ENSURE The estimated minimum and its corresponding location $(\bm{x}^*, f_{\text{min}})$.		
		\STATE \textbf{Design the experiment:} 
		Generate $N_{\text{init}}$ initial sample points $\mathbf{X} = \{\bm{x}^{(i)}\}_{i=1}^{N_{\text{init}}}$, evaluate the objective function
		$\mathbf{Y} = \{f(\bm{x}^{(i)})\}_{i=1}^{N_{\text{init}}}$
		and initialize iteration counter $n = N_{\text{init}}$.
		\STATE \textbf{Set initial best solution:} 
		Set \(\bm{x}^* =\underset{1 \leq i \leq n}{\arg\min}\;f(\bm{x}^{(i)})\) and  \(f_{\text{min}} = f(\bm{x}^*)\).
		\WHILE{$n < N_{\text{max}}$}
		\STATE \textbf{Train the GP:} 
		Fit the GP model using the current dataset.
		\STATE \textbf{Select the next infill point:} 
		Find the point that maximizes the acquisition function
		\[
		\bm{x}_{\text{next}} = \arg \max_{\bm{x} \in \mathcal{X}} \text{EI}(\bm{x}).
		\]
		\STATE \textbf{Evaluate and update dataset:} 
		Compute the objective function at $\bm{x}_{\text{next}}$ and update the dataset
		$
		\mathbf{X} = \mathbf{X} \cup \{\bm{x}_{\text{next}}\}, \quad \mathbf{Y} = \mathbf{Y} \cup \{f(\bm{x}_{\text{next}})\}.
		$
		\STATE \textbf{Update current minimum:}  
		Set \(\bm{x}^* =\underset{1 \leq i \leq n}{\arg\min}\;f(\bm{x}^{(i)})\) , \(f_{\text{min}} = f(\bm{x}^*)\) , and increment \( n = n + 1 \).
		
		\ENDWHILE
		
	\end{algorithmic}	
\end{algorithm}

Initially, a set of \( N_{\text{init}} \) sample points is selected, and their corresponding function values are evaluated. In each iteration, BO constructs a GP model using the current dataset to approximate the objective function. It then selects the next evaluation point by maximizing the EI acquisition function, balancing exploration and exploitation. The objective function is evaluated at the selected point, and the dataset is updated with the new point. This process repeats until the maximum number of evaluations is reached. The computational framework of BO is summarized in Algorithm~\ref{algorithm_BO}.

BO is particularly well-suited for problems with low-dimensional search spaces. In low dimensions, acquisition functions like EI are relatively easy to optimize. As the dimension increases, it becomes more and more difficult to optimize the acquisition function as the search space increases exponentially with the dimension. Therefore, one of the major challenges to extend BO to high dimensions is to tackle the high dimensionality of the acquisition function.

\section{Proposed Approach}
\label{section_coordinate_descent}
In this work, we propose the Adaptive Dropout approach (AdaDropout) to address this problem. Instead of selecting all acquisition points in the original high-dimensional space, the proposed AdaDropout approach dynamically reduces the dimension of the acquisition point to ease the difficulty of optimizing the acquisition function. The details of the proposed algorithm are outlined below.

\subsection{The Basic Idea}

The core idea of AdaDropout is to adaptively adjust the dimensionality of the acquisition function. Initially, the algorithm explores the entire \(D\)-dimensional space using a standard acquisition function (e.g., Expected Improvement, EI). When a newly sampled point does not improve upon the current best solution, it suggests that the high-dimensional search may be inefficient. In this case, we drop one variable, reducing the number of optimizing dimensions from \(d\) to \(d-1\) (with \(d = D\) initially). This can be formally expressed as:
\[
d \leftarrow 
\begin{cases}
	d - 1, & \text{if } f(\bm{x}_{\text{next}}) > f_{\min} \text{ and } d > 1, \\
	d, & \text{otherwise},
\end{cases}
\]
where \( f(\bm{x}_{\text{next}}) \) is the function value at the candidate point and \( f_{\min} \) is the current best (minimum) value.

This adaptive dropout strategy gradually concentrates the search onto promising subspaces. When improvement is detected (i.e., \( f(\bm{x}_{\text{next}}) < f_{\min} \)), the best solution is updated and the current dimensionality \( d \) is maintained, ensuring that the search remains sufficiently explorative. Over iterations, the search space transitions from a full \(D\)-dimensional global search to a lower-dimensional local search, mitigating the curse of dimensionality.

\subsection{The Computational Framework}

\begin{algorithm}
	\caption{Computational Framework of the Proposed AdaDropout Algorithm}
	\label{algorithm_ADR_BO}
	\begin{algorithmic}[1]
		\REQUIRE $N_{\text{init}}$: number of initial samples; 
		$N_{\text{max}}$: maximum allowed function evaluations; 
		$D$: dimensionality of the search space.
		\ENSURE The estimated minimum and its corresponding location $(\bm{x}^*, f_{\text{min}})$.
		
		\STATE \textbf{Design the experiment:} 
		Generate $N_{\text{init}}$ initial sample points $\mathbf{X} = \{\bm{x}^{(i)}\}_{i=1}^{N_{\text{init}}}$, evaluate the objective function
		$\mathbf{Y} = \{f(\bm{x}^{(i)})\}_{i=1}^{N_{\text{init}}}$
		, initialize iteration counter $n = N_{\text{init}}$ and set optimizing dimensionality $d = D$.
		\STATE \textbf{Set initial best solution:} 
		Set \(\bm{x}^* =\underset{1 \leq i \leq n}{\arg\min}\;f(\bm{x}^{(i)})\) and  \(f_{\text{min}} = f(\bm{x}^*)\).

		\WHILE{$n < N_{\text{max}}$}
		\STATE \textbf{Select optimizing variables:} 
		Randomly select a subspace of $d$ dimensions from the search space for optimization.
		
		\STATE \textbf{Train the GP:} 
		Fit the GP model using the current dataset.
		
		\STATE \textbf{Select the next infill point:} 
		Maximize the acquisition function over the selected variables
		\[
		\bm{y}_\text{next} = \underset{\bm{y} \in \mathcal{Y}}{\arg \max} \; \text{ESSI}_d(\bm{y})
		\]
		Form the new solution
		\[
		\bm{x}_\text{next} = [x_{\text{min},1}, \dots, y_{\text{next},1}, \dots, x_{\text{min},i}, \dots, y_{\text{next},d}, \dots, x_{\text{min},D}].
		\]
		
		\STATE \textbf{Evaluate and update dataset:} 
		Compute the objective function at $\bm{x}_{\text{next}}$ and update the dataset
		\[
		\mathbf{X} = \mathbf{X} \cup \{\bm{x}_{\text{next}}\}, \quad \mathbf{Y} = \mathbf{Y} \cup \{f(\bm{x}_{\text{next}})\}.
		\]
		
		\STATE \textbf{Update dimensionality:} 
		If \( f(\bm{x}_{\text{next}}) > f_{\text{min}} \) and \( d > 1 \), reduce the number of optimizing variables by setting
		\[
		d = d - 1.
		\]
		
		\STATE \textbf{Update current minimum:} 
		Set \(\bm{x}^* =\underset{1 \leq i \leq n}{\arg\min}\;f(\bm{x}^{(i)})\) ,  \(f_{\text{min}} = f(\bm{x}^*)\) , and increment \( n = n + 1 \).

		\ENDWHILE
	\end{algorithmic}
\end{algorithm}

\begin{figure}
	\centering
	\includegraphics[width=0.9\linewidth]{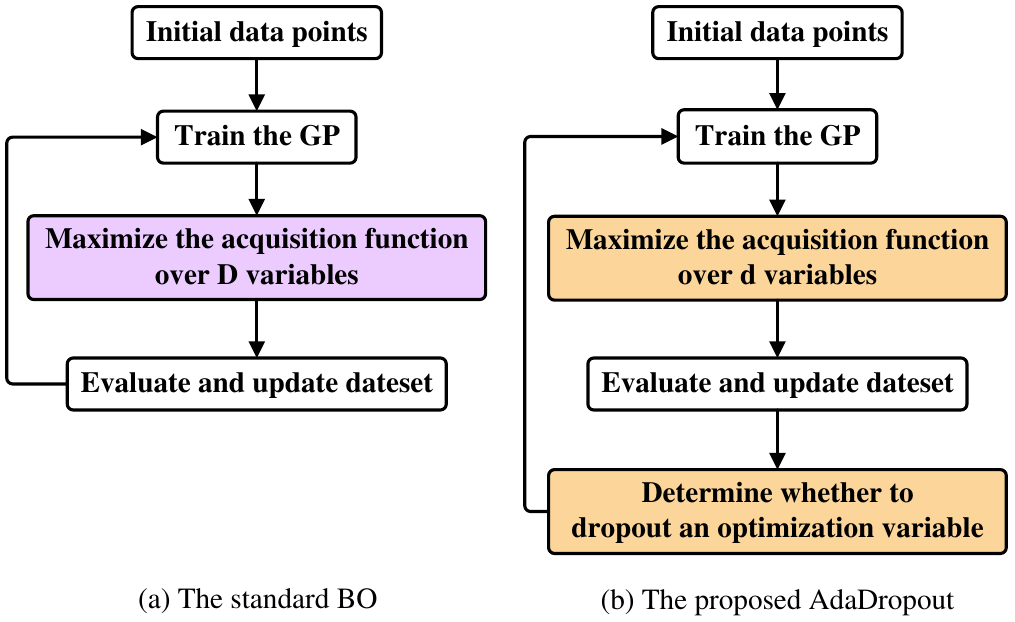}
	\caption{Main Steps of the Standard BO and the Proposed AdaDropout}
	\label{flowchart}
\end{figure}

The computational framework of AdaDropout is outlined in Algorithm~\ref{algorithm_ADR_BO}.

\begin{enumerate} 
	\item \emph{Design the experiment:} In the initialization phase, $N_{\text{init}}$ initial samples are generated using a design of experiment (DoE) method, such as Latin hypercube sampling. These initial samples are evaluated with the expensive objective function, and the results are used to populate the initial dataset. And we set initial the number of optimizing variables \( d \) as \( D \).
	
	\item \emph{Set initial best solution:} The best solution among the initial samples is identified as \( (\bm{x}^\star, f_\text{min}) \), where \( \bm{x}^\star \) is the location of the current minimum, and \( f_\text{min} \) is its corresponding value.
	
	\item \emph{Select optimizing variables:} Randomly select $d$ optimizing variables from the original $D$-dimensional search space.
	
	\item \emph{Train the GP:} The GP model is trained using the current dataset \((\mathbf{X},\mathbf{Y})\).
	
	\item \emph{Select the next infill point:} Using the trained GP model, the Expected Subspace Improvement (ESSI) function \cite{zhan_2024_b} is optimized over the selected optimizing variables to calculate the next query point. The input subspace is defined as \(\mathcal{Y} \subseteq \mathbb{R}^d\). The corresponding dimensions in \( \bm{x}^\star \) are replaced with the optimized values, forming the new candidate solution \( \bm{x}_\text{next} \).
	
	\item \emph{Evaluate and update dataset:} The objective function is evaluated at the new candidate point \( \bm{x}_{\text{next}} \), yielding \( f(\bm{x}_{\text{next}}) \). This point and its function value are added to the dataset.
	
	\item \emph{Update dimensionality:} If the function value at the newly sampled point is worse than the current best and there is more than one variable being optimized, the algorithm reduces the number of optimizing variables by one.
	
	\item \emph{Update current minimum:} The algorithm updates the best-known solution and its corresponding function value. Then, the iteration counter is incremented, and the process repeats until the maximum number of function evaluations is reached.
\end{enumerate}

The flowcharts of the standard BO and the proposed AdaDropout are illustrated in Fig.~\ref{flowchart}. It can be observed that the key differences between these two algorithms lie in the acquisition function optimization process and the dimensionality management. In each iteration, the standard BO maximizes the acquisition function over the entire $D$-dimensional space, which leads to significant computational cost and optimization difficulty in high-dimensional problems. In contrast, AdaDropout maximizes the acquisition function over a reduced $d$-dimensional subproblem. Additionally, after each iteration, AdaDropout determines whether to further drop variable based on the optimization progress. Since the reduced dimension $d$ is much smaller than the original dimension $D$, AdaDropout effectively alleviates the computational burden and difficulty of high-dimensional optimization by dynamically adjusting the optimization subspace.

\subsection{Illustrative Example}
\label{sec:example}

\begin{figure}[htbp]
	\centering
	\includegraphics[width=0.57\linewidth]{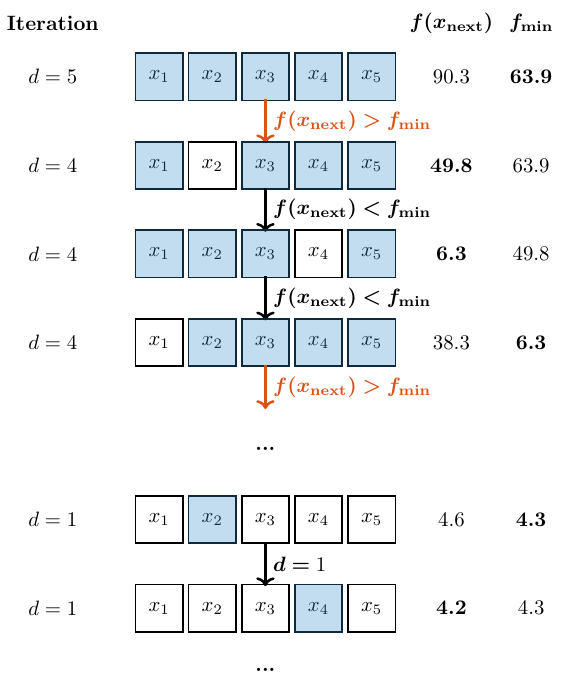}
	\caption{Illustration of the AdaDropout strategy. \textbf{Left:} The current dimensionality of the acquisition function. \textbf{Middle:} A grid showing the selected coordinates (blue squares) for the current iteration. \textbf{Right:} Comparison between \( f_{\min} \) and \( f(x_{\text{next}}) \) used to decide whether to drop an optimization variable.}
	\label{fig_ADR}
\end{figure}

Fig.~\ref{fig_ADR} visually demonstrates the AdaDropout strategy using a five-dimensional optimization problem. The process is as follows:

\begin{itemize}
	\item \textbf{Initial Global Search:} The algorithm starts in the full five-dimensional space. The first acquisition point is selected using the EI criterion, and its function value \( f(\bm{x}_{\text{next}}) = 90.3 \) is compared with the current best \( f_{\min} = 63.9 \).
	\item \textbf{Dimensionality Reduction:} Since \( f(\bm{x}_{\text{next}}) > f_{\min} \), the current high-dimensional search does not yield an improvement. Thus, one variable is dropped, reducing the search space to four dimensions.
	\item \textbf{Focused Local Search:} In the subsequent iteration, the ESSI criterion is used over the four-dimensional subspace, yielding a new point with \( f(\bm{x}_{\text{next}}) = 49.8 \). Because \( f(\bm{x}_{\text{next}}) < f_{\min} \), the best solution is updated and the search continues in the four-dimensional space.
	\item \textbf{Continued Adaptation:} This process repeats, with the algorithm reducing the dimensionality when no improvement is observed, until eventually only one optimization variable remains.
\end{itemize}

In summary, the AdaDropout algorithm intelligently blends global exploration and local exploitation by gradually reducing the dimensionality of the optimization subspace. This adaptive strategy not only mitigates the challenges posed by high-dimensional optimization but also enhances the efficiency and effectiveness of the Bayesian Optimization process.

\section{Numerical Experiments}
\label{section_experiment}
In this section, we conduct numerical experiments to evaluate the performance of the proposed Adaptive Dropout approach (AdaDropout). We compare AdaDropout with the standard Bayesian optimization and six state-of-the-art HDBO algorithms to evaluate its performances in handling high-dimensional optimization problems. The MATLAB implementation of the proposed AdaDropout used in these experiments is publicly available at \url{https://github.com/huang-jundi/Adaptive_Dropout}.

\subsection{Experiment Settings}
The settings for the numerical experiments are provided below.
\begin{enumerate} 
	\item \emph{Test Problem:} We evaluate the performance of the AdaDropout algorithm using benchmark functions from the CEC 2013 test suite~\cite{liang_2013} and the CEC 2017 test suite~\cite{Awad_2017}.
	The CEC 2013 test suite consists of 28 problems, where \( f_\text{1} \) to \( f_\text{5} \) are unimodal problems, \( f_\text{6} \) to \( f_\text{20} \) are multimodal problems, and \( f_\text{21} \) to \( f_\text{28} \) are composition problems. The CEC 2017 test suite includes 29 problems, with \( f_\text{1} \) and \( f_\text{3} \) as unimodal problems, \( f_\text{4} \) to \( f_\text{10} \) as simple multimodal problems, \( f_\text{11} \) to \( f_\text{20} \) as hybrid problems, and \( f_\text{21} \) to \( f_\text{30} \) as composition problems.
	
	\item \emph{Experimental Design:} The dimensionality of the optimization problems is set to 100. To initialize the optimization process, we generate the initial samples using Latin Hypercube Sampling (LHS), with the number of samples fixed at 200. The maximum number of function evaluations for each experiment is limited to 1000.
	
	\item \emph{GP Model:} For constructing the Gaussian Process (GP) surrogate model, a radial basis function kernel is employed. The range of the kernel hyperparameters is constrained within \([0.01, 100]\), and the model is trained using the standard maximum likelihood estimation approach.
	
	\item \emph{Infill Selection:} The ESSI acquisition function is optimized using a Genetic Algorithm (GA). The settings for GA differ depending on the algorithm. For the standard BO approach, the GA population size is set to 200, with 100 generations. For AdaDropout, the GA population size is \(\max(10, 4d)\), where \(d\) is the number of optimizing variables, and the number of generations is set to \(200d / \text{population size}\).
	
	\item \emph{Number of Runs:} Each algorithm is independently run 30 times with different initial samples. To ensure a fair comparison, the same set of initial samples is used across all algorithms in each experiment.
	
	\item \emph{Experiment Environment:} The experiments are conducted on a system running Windows 10, equipped with an Intel Core i9-10900X CPU and 64 GB of RAM. This computational setup ensures a consistent environment for evaluating the algorithms.
	
\end{enumerate}

The proposed AdaDropout is compared with standard BO and six state-of-the-art high-dimensional Bayesian optimization (HDBO) algorithms, including Add-GP-UCB~\cite{Kandasamy_2015}, Dropout~\cite{Li_2017}, CoordinateLineBO~\cite{Kirschner_2019}, TuRBO~\cite{Eriksson_2019}, MCTS-VS~\cite{Song_2022} and TAS-BO~\cite{GUI_2024}. Each algorithm adopts a unique strategy to tackle the challenges of high-dimensional optimization. The descriptions of the six state-of-the-art algorithms are presented below.
\begin{enumerate} 
	\item \emph{Add-GP-UCB:} The Add-GP-UCB~\cite{Kandasamy_2015} algorithm assumes an additive structure, decomposing the objective function into multiple low-dimensional sub-functions. It performs Bayesian optimization within these subspaces, reducing complexity and improving sample efficiency. However, its effectiveness depends on the correctness of the decomposition, and performance may degrade if the function is not strictly additive. The low-dimensional space is set to five in the experiments.
	
	\item \emph{Dropout:} The Dropout~\cite{Li_2017} approach addresses high-dimensional optimization by randomly selecting a subset of variables at each iteration and optimizing only within that subset, thereby reducing computational costs while maintaining exploration. This method avoids restrictive assumptions about the function’s structure and enables optimization in a lower-dimensional subspace. The dimension of the subspace is also set to five in the experiments.
	
	\item \emph{CoordinateLineBO:} The CoordinateLineBO~\cite{Kirschner_2019} optimizes one variable at a time by iterating through coordinates in a random order. We use a genetic algorithm with population size of 10 and 20 generations to optimize the selected coordinate in the experiments.
	
	\item \emph{TuRBO:} The TuRBO~\cite{Eriksson_2019} integrates a trust region strategy into the Bayesian optimization framework, enhancing its local search capabilities. By dynamically adjusting the size and location of trust regions, TuRBO balances global exploration and local exploitation, making it particularly effective for high-dimensional problems with complex landscapes. The number of trust regions is set to five and the number of batch evaluations is set to ten in the experiment.
	
	\item \emph{MCTS-VS:} The MCTS-VS~\cite{Song_2022} employs a Monte Carlo tree search method to partition variables into two categories: important and unimportant. By focusing optimization efforts on the important variables, MCTS-VS reduces the effective dimensionality of the search space, enabling more efficient exploration.
	
	\item \emph{TAS-BO:} The TAS-BO~\cite{GUI_2024} refines optimization by first identifying a promising candidate using a global Gaussian Process and then training a local GP around it for further refinement. This adaptive approach balances global exploration and local exploitation, improving convergence in high-dimensional problems.
	
\end{enumerate}

\subsection{Results on CEC 2013}

To evaluate the statistical significance of the results, we perform a Wilcoxon signed-rank test at a significance level of \(\alpha = 0.05\). In the tables, the symbols \(+\), \(-\), and \(\approx\) indicate that the results of the proposed AdaDropout are significantly better than, worse than, or similar to the results of the compared algorithms, respectively.

Table~\ref{table_HDBOs_CEC2013_100D} reports the average performance of AdaDropout and seven BO algorithms on 100-dimensional CEC 2013 benchmark problems. In these tables, lower objective function values indicate better optimization performance. As evident from the results, AdaDropout achieves a significant advantage over the compared methods.

\begin{table*}
	\renewcommand{\arraystretch}{1.0}
	\caption{Average results obtained by the compared BO algorithms on 100-D CEC 2013 test problems}
	\label{table_HDBOs_CEC2013_100D}
	\centering
	\resizebox{\textwidth}{!}{
		\begin{tabular}{c c c c c c c  c c } 
			\toprule
			$f$  & BO &  Add-GP-UCB & Dropout  &  \makecell[c]{Coordinate\\LineBO} &  TuRBO & MCTS-VS & TAS-BO &  AdaDropout   \\
			\midrule
			$f_{1}$ & 1.55E+03 $+$ & 2.60E+05 $+$ & 7.06E+04 $+$ & 1.29E+03 $+$ & 5.22E+04 $+$ & 9.13E+04 $+$ & -5.03E+02 $+$ & -1.24E+03 \\
			$f_{2}$ & 1.50E+09 $+$ & 1.05E+10 $+$ & 1.14E+09 $+$ & 6.19E+08 $+$ & 2.83E+09 $+$ & 2.79E+09 $+$ & 6.72E+08 $+$ & 3.31E+08 \\
			$f_{3}$ & 2.28E+24 $+$ & 2.37E+24 $+$ & -7.88E+30 $-$ & 7.60E+18 $\approx$ & 3.09E+18 $\approx$ & 2.96E+19 $\approx$ & 4.33E+23 $+$ & 6.06E+18 \\
			$f_{4}$ & 4.82E+05 $\approx$ & 5.16E+05 $\approx$ & 4.80E+05 $\approx$ & 5.31E+05 $+$ & 4.35E+05 $-$ & 5.04E+05 $\approx$ & 4.37E+05 $\approx$ & 4.81E+05 \\
			$f_{5}$ & 4.20E+04 $+$ & 1.26E+05 $+$ & 1.26E+04 $-$ & 2.97E+04 $+$ & 2.86E+04 $+$ & 4.02E+04 $+$ & 1.74E+04 $-$ & 2.28E+04 \\
			$f_{6}$ & 2.62E+03 $+$ & 7.61E+04 $+$ & 9.51E+03 $+$ & 9.50E+02 $+$ & 8.15E+03 $+$ & 1.17E+04 $+$ & 5.30E+01 $\approx$ & 8.82E+00 \\
			$f_{7}$ & 5.63E+08 $+$ & 9.56E+08 $+$ & 8.67E+05 $\approx$ & 1.97E+06 $\approx$ & 2.42E+06 $\approx$ & 6.83E+05 $\approx$ & 1.53E+08 $+$ & 1.78E+06 \\
			$f_{8}$ & -6.79E+02 $\approx$ & -6.79E+02 $\approx$ & -6.79E+02 $\approx$ & -6.79E+02 $\approx$ & -6.79E+02 $\approx$ & -6.79E+02 $\approx$ & -6.79E+02 $\approx$ & -6.79E+02 \\
			$f_{9}$ & -4.27E+02 $+$ & -4.26E+02 $+$ & -4.43E+02 $-$ & -4.37E+02 $\approx$ & -4.52E+02 $-$ & -4.25E+02 $+$ & -4.45E+02 $\approx$ & -4.36E+02 \\
			$f_{10}$ & 3.65E+03 $+$ & 4.29E+04 $+$ & 8.21E+03 $+$ & 2.01E+03 $+$ & 8.21E+03 $+$ & 1.19E+04 $+$ & 5.25E+02 $\approx$ & 5.87E+02 \\
			$f_{11}$ & 1.20E+03 $+$ & 4.45E+03 $+$ & 1.12E+03 $+$ & 7.58E+02 $+$ & 1.49E+03 $+$ & 1.82E+03 $+$ & 8.08E+02 $+$ & 6.05E+02 \\
			$f_{12}$ & 1.25E+03 $-$ & 4.56E+03 $+$ & 2.50E+03 $+$ & 2.26E+03 $+$ & 1.93E+03 $+$ & 1.77E+03 $+$ & 9.10E+02 $-$ & 1.45E+03 \\
			$f_{13}$ & 1.38E+03 $-$ & 4.49E+03 $+$ & 2.70E+03 $+$ & 2.61E+03 $+$ & 2.01E+03 $+$ & 1.84E+03 $+$ & 9.96E+02 $-$ & 1.62E+03 \\
			$f_{14}$ & 3.44E+04 $+$ & 3.19E+04 $+$ & 1.83E+04 $+$ & 1.43E+04 $-$ & 2.32E+04 $+$ & 1.91E+04 $+$ & 3.45E+04 $+$ & 1.59E+04 \\
			$f_{15}$ & 3.41E+04 $+$ & 3.43E+04 $+$ & 3.28E+04 $+$ & 2.56E+04 $\approx$ & 3.15E+04 $+$ & 3.41E+04 $+$ & 3.41E+04 $+$ & 2.49E+04 \\
			$f_{16}$ & 2.06E+02 $\approx$ & 2.06E+02 $\approx$ & 2.06E+02 $\approx$ & 2.06E+02 $\approx$ & 2.06E+02 $\approx$ & 2.06E+02 $\approx$ & 2.06E+02 $\approx$ & 2.06E+02 \\
			$f_{17}$ & 1.88E+03 $+$ & 9.13E+03 $+$ & 2.92E+03 $+$ & 1.36E+03 $\approx$ & 3.20E+03 $+$ & 4.33E+03 $+$ & 1.52E+03 $+$ & 1.22E+03 \\
			$f_{18}$ & 1.98E+03 $-$ & 9.33E+03 $+$ & 4.48E+03 $+$ & 4.37E+03 $+$ & 3.33E+03 $+$ & 4.30E+03 $+$ & 1.61E+03 $-$ & 2.37E+03 \\
			$f_{19}$ & 9.00E+05 $+$ & 3.58E+07 $+$ & 7.35E+05 $+$ & 1.94E+05 $+$ & 2.68E+06 $+$ & 3.83E+06 $+$ & 8.86E+04 $+$ & 3.10E+04 \\
			$f_{20}$ & 6.50E+02 $\approx$ & 6.50E+02 $\approx$ & 6.50E+02 $\approx$ & 6.50E+02 $\approx$ & 6.50E+02 $\approx$ & 6.50E+02 $\approx$ & 6.50E+02 $\approx$ & 6.50E+02 \\
			$f_{21}$ & 1.09E+04 $+$ & 2.29E+04 $+$ & 2.30E+02 $\approx$ & 1.34E+04 $+$ & 1.32E+04 $+$ & 1.18E+04 $+$ & 1.07E+04 $+$ & 9.83E+03 \\
			$f_{22}$ & 3.74E+04 $+$ & 3.60E+04 $+$ & 2.14E+04 $+$ & 1.82E+04 $-$ & 2.51E+04 $+$ & 2.29E+04 $+$ & 3.72E+04 $+$ & 1.99E+04 \\
			$f_{23}$ & 3.70E+04 $+$ & 3.71E+04 $+$ & 3.54E+04 $+$ & 3.02E+04 $\approx$ & 3.46E+04 $+$ & 3.68E+04 $+$ & 3.70E+04 $+$ & 3.07E+04 \\
			$f_{24}$ & 1.66E+03 $+$ & 2.65E+03 $+$ & 1.77E+03 $+$ & 1.65E+03 $+$ & 1.68E+03 $+$ & 1.65E+03 $+$ & 1.54E+03 $-$ & 1.61E+03 \\
			$f_{25}$ & 1.80E+03 $\approx$ & 2.23E+03 $+$ & 1.99E+03 $+$ & 1.84E+03 $+$ & 1.91E+03 $+$ & 1.83E+03 $+$ & 1.80E+03 $\approx$ & 1.80E+03 \\
			$f_{26}$ & 1.97E+03 $+$ & 2.03E+03 $+$ & 1.92E+03 $\approx$ & 1.91E+03 $\approx$ & 1.94E+03 $+$ & 1.94E+03 $+$ & 1.95E+03 $+$ & 1.91E+03 \\
			$f_{27}$ & 6.03E+03 $+$ & 7.28E+03 $+$ & 6.21E+03 $+$ & 5.94E+03 $+$ & 6.36E+03 $+$ & 5.96E+03 $+$ & 4.80E+03 $-$ & 5.66E+03 \\
			$f_{28}$ & 3.49E+04 $+$ & 3.73E+04 $+$ & 1.83E+04 $-$ & 2.10E+04 $\approx$ & 2.17E+04 $\approx$ & 2.14E+04 $\approx$ & 2.85E+04 $+$ & 2.13E+04 \\
			\midrule   
			{$+$/$\approx$/$-$ }  &   20/5/3  &  24/4/0  &  17/7/4  &  15/11/2   & 20/6/2  &  21/7/0  &  14/8/6  & N.A. \\
			\bottomrule
		\end{tabular}
	}
\end{table*}

\begin{figure*}
	\centering
	\begin{minipage}{\textwidth}
		\subfloat[$f_\text{1}$]{\includegraphics[width=0.5\linewidth]{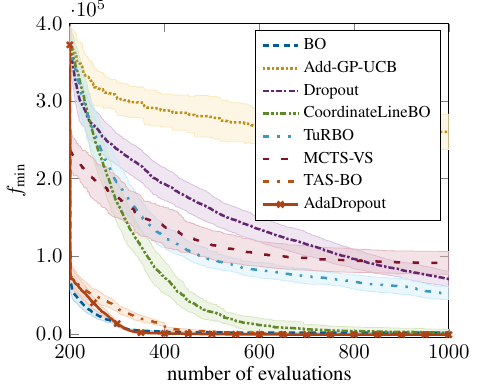}}\hfill
		\subfloat[$f_\text{2}$]{\includegraphics[width=0.5\linewidth]{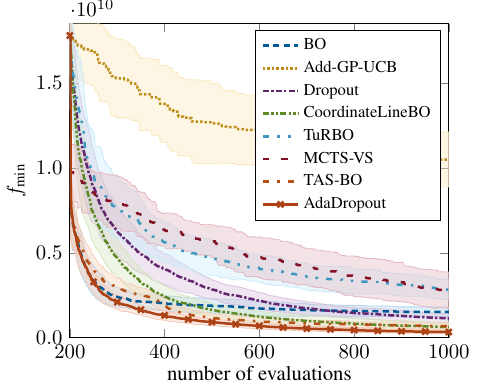}}
	\end{minipage}
	\\
	\begin{minipage}{\textwidth}
		\subfloat[$f_\text{6}$]{\includegraphics[width=0.5\linewidth]{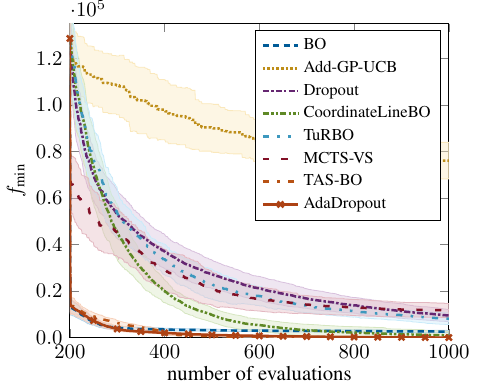}}\hfill		
		\subfloat[$f_\text{11}$]{\includegraphics[width=0.5\linewidth]{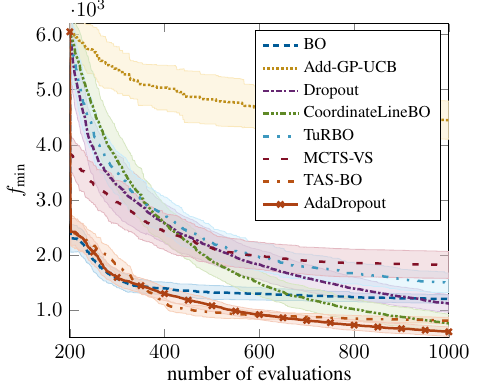}}
	\end{minipage}
	\\
	\begin{minipage}{\textwidth}
		\subfloat[$f_\text{17}$]{\includegraphics[width=0.5\linewidth]{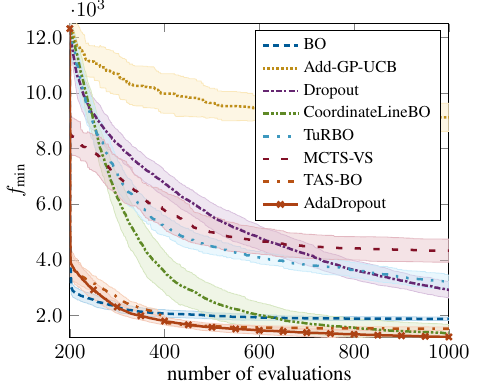}}\hfill
		\subfloat[$f_\text{19}$]{\includegraphics[width=0.5\linewidth]{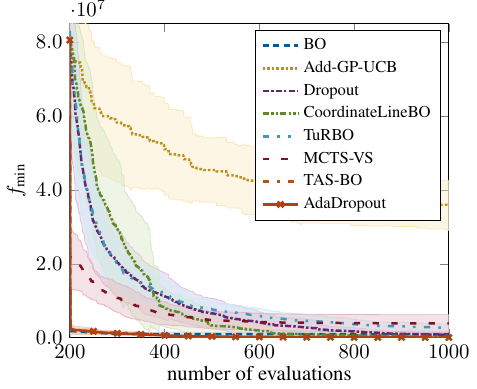}}
	\end{minipage}
	\caption{Convergence histories of the BOs and the proposed AdaDropout on the CEC2013 test problems. (a) $f_\text{1}$. (b) $f_\text{2}$. (c) $f_\text{6}$. (d) $f_\text{11}$. (e) $f_\text{17}$. (f) $f_\text{19}$.}
	\label{fig_comparison_CEC2013_HDBO}
\end{figure*}

Standard BO optimizes the acquisition function in the full $D$-dimensional space throughout the optimization process, which may lead to inefficient acquisition function optimization in high-dimensional space. In contrast, the proposed AdaDropout dynamically adjusts the optimization dimensionality, allowing for efficient exploration in high-dimensional space and refined exploitation in lower dimensions. On the CEC 2013 benchmark problems, AdaDropout outperforms standard BO on twenty functions, achieves comparable performance on five, and falls behind on three.

Add-GP-UCB assumes an additive structure in the objective function and optimizes within separate subspaces. While effective in problems with strong additive decompositions, its performance degrades when the true objective function lacks this structure. AdaDropout, on the other hand, does not rely on such structural assumptions and adapts its optimization space dynamically. On the CEC 2013 benchmark problems, AdaDropout outperforms Add-GP-UCB on twenty-four functions and achieves comparable performance on four.

Dropout employs a fixed-dimensional optimization strategy by randomly selecting a subset of variables for optimization in each iteration, while keeping the remaining variables fixed. In contrast, AdaDropout enhances this approach by adaptively reducing the optimization dimensionality only when no improvement is observed in the objective function. On the CEC 2013 benchmark problems, AdaDropout outperforms Dropout on seventeen functions, achieves comparable performance on seven, and falls behind on four. 

CoordinateLineBO employs a coordinate-wise optimization strategy, optimizing one dimension at a time while keeping others fixed. While this approach mitigates the curse of dimensionality, it may struggle in complex optimization landscapes where interactions between variables are significant. In contrast, AdaDropout retains a more flexible dimensionality reduction strategy, maintaining higher exploration efficiency in the early optimization phase while refining solutions in the later stage. On the CEC 2013 benchmark problems, AdaDropout outperforms CoordinateLineBO on fifteen functions, achieves comparable performance on eleven, and falls behind on two.

TuRBO employs multiple local trust regions for optimization, effectively handling high-dimensional spaces but potentially limiting global exploration.  In contrast, AdaDropout adaptively adjusts the optimization dimensionality, striking a balance between exploration and exploitation. This leads to superior performance. On the CEC 2013 benchmark problems, AdaDropout outperforms TuRBO on twenty functions, achieves comparable performance on six, and falls behind on two.

MCTS-VS employs a Monte Carlo tree search method to partition variables into important and unimportant categories, focusing optimization efforts on the important ones to effectively reduce the search space dimensionality. However, it may struggle with high-dimensional problems where variable interactions are complex. On the CEC 2013 benchmark problems, AdaDropout outperforms MCTS-VS on twenty-one functions, achieves comparable performance on seven, and falls behind on none.

TAS-BO employs a two-step optimization strategy, first identifying a candidate point using a global GP model and then refining it with a local GP model before evaluation. This approach enhances local search capability, improving BO’s effectiveness on high-dimensional problems. However, TAS-BO still evaluates points in the full D-dimensional space, whereas AdaDropout dynamically reduces the optimization dimensionality, further improving efficiency. On the CEC 2013 benchmark problems, AdaDropout outperforms TAS-BO on fourteen functions, achieves comparable performance on eight, and falls behind on six.

The convergence histories of the compared algorithms on six selected functions \( f1 \), \( f2 \), \( f6 \), \( f11 \), \( f17 \), and \( f19 \) from the CEC 2013 test suite are presented in Fig.~\ref{fig_comparison_CEC2013_HDBO}, where the average performance over 30 independent runs is plotted. From the convergence curves, it is evident that AdaDropout demonstrates superior performance across the majority of the test problems.

For instance, in the cases of \(f_2\), \(f_6\), and \(f_{11}\), AdaDropout exhibits a steep decline in \(f_{\text{min}}\) during the early iterations, followed by a stable and consistent improvement. This suggests that AdaDropout effectively exploits the initial high-dimensional optimization phase for rapid performance gains while maintaining steady progress in the later stages, ultimately outperforming other algorithms.  

Compared with standard Bayesian optimization, AdaDropout initially performs slightly worse, which may be attributed to the gradual reduction in the dimensionality of the acquisition function during the early iterations. However, in the later stages, AdaDropout exhibits steady improvements, whereas standard BO suffers from the curse of high-dimensional optimization, leading to performance stagnation. This adaptive dimension reduction mechanism allows AdaDropout to achieve superior overall performance on many benchmark problems.

Compared with other state-of-the-art methods, AdaDropout shows a clear advantage in the early optimization phase, as evident from the rapid descent in the convergence curves. This advantage arises from our adaptive dropout mechanism, which starts by optimizing in the full \(D\)-dimensional space and gradually reduces the number of optimizing variables when progress slows. By iteratively dropping less informative variables and refining the search in lower-dimensional subspaces, AdaDropout mitigates the challenges of high-dimensional optimization and sustains steady improvements throughout the optimization process.  

On most benchmark problems, AdaDropout achieves highly competitive results across different optimization phases. Its ability to balance early-stage exploration with late-stage exploitation underscores its adaptability and precision in identifying optimal solutions.

\begin{table*}
	\renewcommand{\arraystretch}{1.0}
	\caption{Average results obtained by the compared BO algorithms on 100-D CEC 2017 test problems}
	\label{table_HDBOs_CEC2017_100D}
	\centering
	\resizebox{\textwidth}{!}{
		\begin{tabular}{c c c c c c c  c c } 
			\toprule
			$f$  & BO &  Add-GP-UCB & Dropout  &  \makecell[c]{Coordinate\\LineBO} &  TuRBO & MCTS-VS & TAS-BO &  AdaDropout   \\
			\midrule
			$f_{1}$ & 1.17E+10 $+$ & 4.03E+11 $+$ & 9.60E+10 $+$ & 3.51E+09 $+$ & 9.69E+10 $+$ & 1.49E+11 $+$ & 3.10E+09 $+$ & 1.29E+09 \\
			$f_{3}$ & 1.52E+06 $+$ & 1.37E+07 $+$ & 9.54E+05 $\approx$ & 9.65E+05 $\approx$ & 3.01E+06 $\approx$ & 1.58E+06 $+$ & 1.87E+06 $+$ & 1.04E+06 \\
			$f_{4}$ & 1.43E+04 $+$ & 1.49E+05 $+$ & 1.67E+04 $+$ & 6.65E+03 $+$ & 4.86E+04 $+$ & 2.45E+04 $+$ & 3.18E+03 $-$ & 4.21E+03 \\
			$f_{5}$ & 1.61E+03 $+$ & 2.53E+03 $+$ & 2.02E+03 $+$ & 1.70E+03 $+$ & 1.76E+03 $+$ & 1.67E+03 $+$ & 1.25E+03 $-$ & 1.32E+03 \\
			$f_{6}$ & 6.63E+02 $+$ & 7.31E+02 $+$ & 6.72E+02 $+$ & 6.40E+02 $+$ & 6.60E+02 $+$ & 6.61E+02 $+$ & 6.49E+02 $+$ & 6.30E+02 \\
			$f_{7}$ & 2.26E+03 $+$ & 9.33E+03 $+$ & 4.15E+03 $+$ & 2.32E+03 $+$ & 3.76E+03 $+$ & 4.68E+03 $+$ & 1.89E+03 $-$ & 2.07E+03 \\
			$f_{8}$ & 1.92E+03 $+$ & 2.92E+03 $+$ & 2.37E+03 $+$ & 2.00E+03 $+$ & 2.08E+03 $+$ & 2.00E+03 $+$ & 1.58E+03 $-$ & 1.64E+03 \\
			$f_{9}$ & 4.45E+04 $+$ & 1.54E+05 $+$ & 1.05E+05 $+$ & 8.00E+04 $+$ & 8.23E+04 $+$ & 5.08E+04 $+$ & 3.78E+04 $+$ & 2.49E+04 \\
			$f_{10}$ & 3.44E+04 $+$ & 3.43E+04 $+$ & 2.60E+04 $+$ & 2.25E+04 $\approx$ & 2.92E+04 $+$ & 3.45E+04 $+$ & 3.47E+04 $+$ & 2.34E+04 \\
			$f_{11}$ & 4.93E+05 $+$ & 4.56E+05 $+$ & 1.83E+05 $-$ & 2.66E+05 $\approx$ & 3.25E+05 $+$ & 4.71E+05 $+$ & 4.06E+05 $+$ & 2.58E+05 \\
			$f_{12}$ & 1.48E+10 $+$ & 1.86E+11 $+$ & 2.36E+10 $+$ & 4.95E+09 $+$ & 3.29E+10 $+$ & 5.66E+10 $+$ & 4.71E+09 $+$ & 1.37E+09 \\
			$f_{13}$ & 4.23E+09 $+$ & 4.03E+10 $+$ & 3.91E+09 $+$ & 1.73E+09 $+$ & 3.36E+09 $+$ & 1.21E+10 $+$ & 1.69E+09 $+$ & 8.82E+08 \\
			$f_{14}$ & 9.31E+07 $\approx$ & 2.83E+08 $+$ & 5.83E+07 $-$ & 7.91E+07 $\approx$ & 8.29E+07 $\approx$ & 1.25E+08 $+$ & 8.01E+07 $\approx$ & 9.03E+07 \\
			$f_{15}$ & 3.36E+09 $+$ & 1.80E+10 $+$ & 9.15E+08 $\approx$ & 1.59E+09 $+$ & 1.28E+09 $+$ & 4.78E+09 $+$ & 1.31E+09 $+$ & 7.27E+08 \\
			$f_{16}$ & 1.28E+04 $+$ & 2.16E+04 $+$ & 9.70E+03 $-$ & 1.13E+04 $+$ & 1.49E+04 $+$ & 1.25E+04 $+$ & 1.10E+04 $\approx$ & 1.09E+04 \\
			$f_{17}$ & 8.20E+05 $+$ & 8.19E+06 $+$ & 1.58E+04 $-$ & 2.43E+06 $+$ & 7.63E+04 $-$ & 4.14E+04 $-$ & 1.41E+05 $-$ & 2.83E+05 \\
			$f_{18}$ & 1.31E+08 $+$ & 5.38E+08 $+$ & 5.65E+07 $-$ & 1.59E+08 $+$ & 1.32E+08 $+$ & 2.37E+08 $+$ & 9.25E+07 $\approx$ & 9.52E+07 \\
			$f_{19}$ & 2.90E+09 $+$ & 1.74E+10 $+$ & 9.65E+08 $\approx$ & 1.29E+09 $+$ & 1.20E+09 $+$ & 5.08E+09 $+$ & 1.40E+09 $+$ & 7.09E+08 \\
			$f_{20}$ & 8.88E+03 $+$ & 8.91E+03 $+$ & 6.22E+03 $-$ & 7.21E+03 $\approx$ & 6.06E+03 $-$ & 8.48E+03 $+$ & 8.86E+03 $+$ & 7.09E+03 \\
			$f_{21}$ & 3.59E+03 $+$ & 4.70E+03 $+$ & 3.90E+03 $+$ & 3.55E+03 $+$ & 4.05E+03 $+$ & 3.57E+03 $+$ & 3.43E+03 $\approx$ & 3.42E+03 \\
			$f_{22}$ & 3.70E+04 $+$ & 3.67E+04 $+$ & 2.87E+04 $+$ & 2.48E+04 $\approx$ & 3.20E+04 $+$ & 3.66E+04 $+$ & 3.68E+04 $+$ & 2.54E+04 \\
			$f_{23}$ & 4.14E+03 $+$ & 6.10E+03 $+$ & 4.51E+03 $+$ & 3.93E+03 $+$ & 4.68E+03 $+$ & 4.21E+03 $+$ & 3.84E+03 $\approx$ & 3.83E+03 \\
			$f_{24}$ & 4.59E+03 $+$ & 9.63E+03 $+$ & 5.96E+03 $+$ & 4.55E+03 $+$ & 5.51E+03 $+$ & 5.21E+03 $+$ & 4.27E+03 $-$ & 4.35E+03 \\
			$f_{25}$ & 1.17E+04 $+$ & 7.39E+04 $+$ & 1.33E+04 $+$ & 7.06E+03 $+$ & 3.13E+04 $+$ & 2.00E+04 $+$ & 4.85E+03 $-$ & 5.37E+03 \\
			$f_{26}$ & 2.22E+04 $+$ & 5.95E+04 $+$ & 2.94E+04 $+$ & 1.96E+04 $+$ & 3.36E+04 $+$ & 2.37E+04 $+$ & 1.75E+04 $-$ & 1.88E+04 \\
			$f_{27}$ & 4.65E+03 $+$ & 1.18E+04 $+$ & 5.39E+03 $+$ & 4.01E+03 $+$ & 5.81E+03 $+$ & 5.26E+03 $+$ & 4.35E+03 $+$ & 3.89E+03 \\
			$f_{28}$ & 1.44E+04 $+$ & 4.95E+04 $+$ & 1.62E+04 $+$ & 1.48E+04 $+$ & 2.32E+04 $+$ & 2.14E+04 $+$ & 5.45E+03 $-$ & 7.46E+03 \\
			$f_{29}$ & 4.98E+05 $+$ & 2.06E+06 $+$ & 1.23E+04 $-$ & 3.13E+05 $+$ & 4.45E+04 $-$ & 1.94E+04 $-$ & 2.09E+05 $+$ & 9.60E+04 \\
			$f_{30}$ & 7.59E+09 $+$ & 3.10E+10 $+$ & 3.18E+09 $+$ & 1.43E+09 $\approx$ & 2.73E+09 $+$ & 7.07E+09 $+$ & 2.48E+09 $+$ & 1.16E+09 \\
			\midrule   
			{$+$/$\approx$/$-$ }  &   28/1/0  &   29/0/0  &  19/3/7 &   22/7/0  &  24/2/3  &  27/0/2 &  15/5/9  & N.A. \\
			\bottomrule
		\end{tabular}
	}
\end{table*}

\begin{figure*}
	\centering
	\begin{minipage}{\textwidth}
		\subfloat[$f_\text{6}$]{\includegraphics[width=0.5\linewidth]{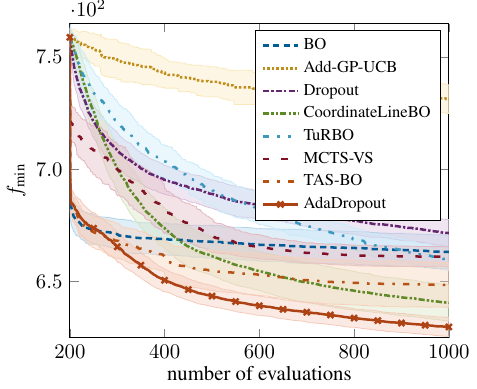}}\hfill
		\subfloat[$f_\text{9}$]{\includegraphics[width=0.5\linewidth]{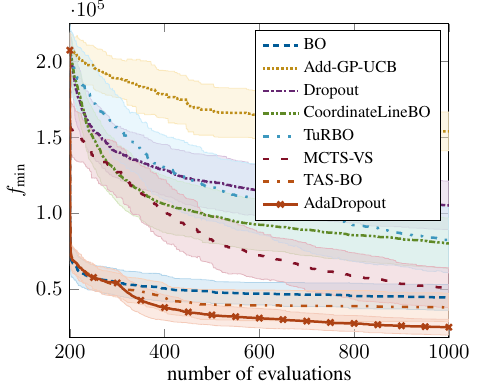}}
	\end{minipage}
	\\
	\begin{minipage}{\textwidth}
		\subfloat[$f_\text{12}$]{\includegraphics[width=0.5\linewidth]{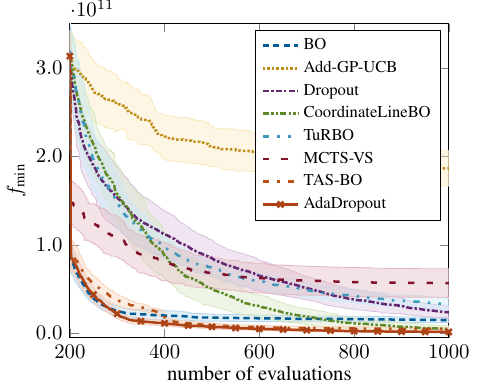}}\hfill		
		\subfloat[$f_\text{13}$]{\includegraphics[width=0.5\linewidth]{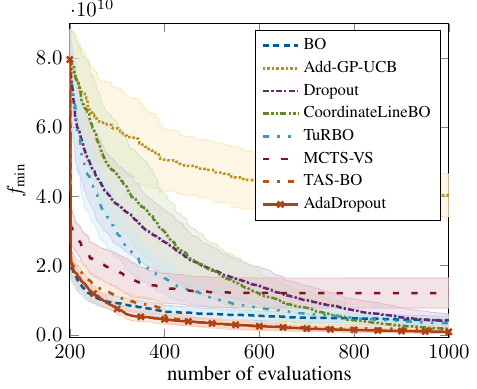}}
	\end{minipage}
	\\
	\begin{minipage}{\textwidth}
		\subfloat[$f_\text{19}$]{\includegraphics[width=0.5\linewidth]{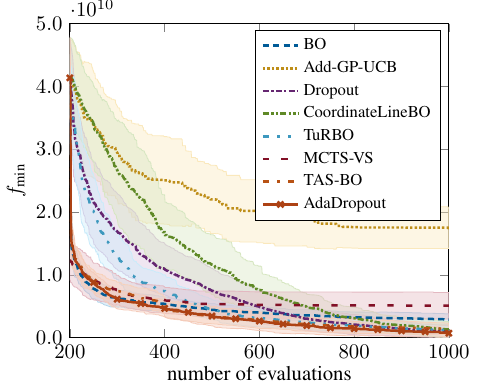}}\hfill
		\subfloat[$f_\text{27}$]{\includegraphics[width=0.5\linewidth]{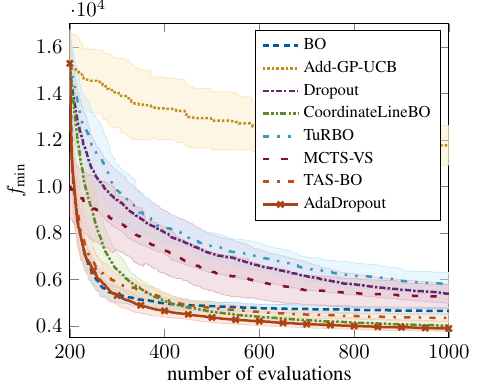}}
	\end{minipage}
	\caption{Convergence histories of the BOs and the proposed AdaDropout on the CEC2017 test problems. (a) $f_\text{6}$. (b) $f_\text{9}$. (c) $f_\text{12}$. (d) $f_\text{13}$. (e) $f_\text{19}$. (f) $f_\text{27}$.}
	\label{fig_comparison_HDBO}
\end{figure*}

\subsection{Results on CEC 2017}

To further evaluate AdaDropout's performance, we test it on 100-dimensional CEC 2017 benchmark problems and present the average optimization results in Table~\ref{table_HDBOs_CEC2017_100D}. Statistical significance is determined using the Wilcoxon signed-rank test at a significance level of \(\alpha = 0.05\). In the tables, the symbols \(+\), \(-\), and \(\approx\) indicate that the performance of the proposed AdaDropout is significantly better than, worse than, or similar to that of the compared methods, respectively.

Table~\ref{table_HDBOs_CEC2017_100D} shows that AdaDropout exhibits strong performance on these high-dimensional problems. Specifically, it outperforms standard BO on twenty-eight benchmark functions. Additionally, it surpasses Add-GP-UCB on twenty-nine functions, Dropout on nineteen functions, CoordinateLineBO on twenty-two functions, TuRBO on twenty-four functions, MCTS-VS on twenty-seven functions, and TAS-BO on fifteen functions, demonstrating its effectiveness against various state-of-the-art methods.

Fig.~\ref{fig_comparison_HDBO} presents the convergence histories of the compared methods on selected CEC 2017 functions. Notably, for functions such as \(f_1\), \(f_6\), \(f_{12}\), and \(f_{24}\), AdaDropout rapidly decreases the objective value in the early iterations, followed by a steady and gradual improvement. This behavior highlights the benefit of the adaptive dimension reduction: an initial global exploration in the full D-dimensional space quickly locates promising regions, after which the search is progressively refined in lower-dimensional subspaces.

In summary, the results on CEC 2017 benchmarks confirm that the adaptive dimensionality reduction mechanism in AdaDropout not only accelerates the initial descent in high-dimensional spaces but also secures sustained improvements in later iterations. This balance between global exploration and local exploitation leads to an overall superior performance compared to standard BO and other state-of-the-art high-dimensional Bayesian optimization methods.

\section{Conclusions}
\label{section_conclusion}

In this work, we present a novel and efficient approach to extend Bayesian optimization to high-dimensional optimization problems. 
The proposed approach is called Adaptive Dropout (AdaDropout) in this work. The proposed AdaDropout adaptively dropout the variables of the acquisition function along the iterations.
The proposed AdaDropout algorithm demonstrates highly competitive performance when compared with the standard BO algorithm and several state-of-the-art high-dimensional Bayesian optimization methods.  
However, our approach does not account for expensive constraints or multiple objectives. Extending the AdaDropout framework to handle constrained optimization and multi-objective problems presents a promising direction for future research and development.




 \bibliographystyle{elsarticle-num-names} 
\bibliography{My_Reference.bib}



\end{document}